\documentclass[runningheads]{llncs}
\usepackage{graphicx}
\usepackage[usenames,dvipsnames,table]{xcolor}
\usepackage{tikz}
\usepackage{comment}
\usepackage{amsmath,amssymb} % define this before the line numbering.
\usepackage{booktabs}
\usepackage{subcaption}
\usepackage{multirow}
\usepackage{float}
\usepackage{wrapfig}

\usepackage[capitalize]{cleveref}
\crefname{section}{Sec.}{Secs.}
\Crefname{section}{Section}{Sections}
\Crefname{table}{Table}{Tables}
\crefname{table}{Tab.}{Tabs.}

\newcommand{\gray}[1]{\textcolor{Gray}{#1}}

\begin{document}
% \renewcommand\thelinenumber{\color[rgb]{0.2,0.5,0.8}\normalfont\sffamily\scriptsize\arabic{linenumber}\color[rgb]{0,0,0}}
% \renewcommand\makeLineNumber {\hss\thelinenumber\ \hspace{6mm} \rlap{\hskip\textwidth\ \hspace{6.5mm}\thelinenumber}}
% \linenumbers
\pagestyle{headings}
\mainmatter
\def\ECCVSubNumber{5318}  % Insert your submission number here

\title{LidarNAS: Unifying and Searching Neural Architectures for 3D Point Clouds
% (Supplementary Material)
} % Replace with your title

% INITIAL SUBMISSION 
\begin{comment}
\titlerunning{ECCV-22 submission ID \ECCVSubNumber} 
\authorrunning{ECCV-22 submission ID \ECCVSubNumber} 
\author{Anonymous ECCV submission}
\institute{Paper ID \ECCVSubNumber}
\end{comment}
%******************

% CAMERA READY SUBMISSION
%\begin{comment}
\titlerunning{LidarNAS}
% If the paper title is too long for the running head, you can set
% an abbreviated paper title here
%
\author{Chenxi Liu \and
Zhaoqi Leng \and
Pei Sun \and
Shuyang Cheng \and
Charles R. Qi \and
Yin Zhou \and
Mingxing Tan \and
Dragomir Anguelov
}
\authorrunning{C. Liu et al.}
% First names are abbreviated in the running head.
% If there are more than two authors, 'et al.' is used.
%
\institute{Waymo LLC \\
\email{\{cxliu, lengzhaoqi, peis, shuyangcheng, rqi, yinzhou, tanmingxing, dragomir\}@waymo.com}
}
%\end{comment}
%******************
\maketitle

%%%%%%%%% ABSTRACT
\begin{abstract}
Developing neural models that accurately understand objects in 3D point clouds is essential for the success of robotics and autonomous driving. However, arguably due to the higher-dimensional nature of the data (as compared to images), existing neural architectures exhibit a large variety in their designs, including but not limited to the views considered, the format of the neural features, and the neural operations used. Lack of a unified framework and interpretation makes it hard to put these designs in perspective, as well as systematically explore new ones. In this paper, we begin by proposing a unified framework of such, with the key idea being factorizing the neural networks into a series of view transforms and neural layers. We demonstrate that this modular framework can reproduce a variety of existing works while allowing a fair comparison of backbone designs. Then, we show how this framework can easily materialize into a concrete neural architecture search (NAS) space, allowing a principled NAS-for-3D exploration. In performing evolutionary NAS on the 3D object detection task on the Waymo Open Dataset, not only do we outperform the state-of-the-art models, but also report the interesting finding that NAS tends to discover the same macro-level architecture concept for both the vehicle and pedestrian classes.
\end{abstract}

%%%%%%%%% BODY TEXT
\section{Introduction}
\label{sec:intro}

Being able to recognize, segment, or detect objects in 3D is one of the fundamental goals of computer vision.
In this paper we consider the point cloud input representation for the wide usage of RGBD cameras in robotics applications, as well as LiDAR sensors in autonomous driving.
There has been a lot of research in this area, including various deep learning based approaches.

But which neural architecture should you choose?
PointNet~\cite{qi2017pointnet}? VoxelNet~\cite{zhou2018voxelnet}? PointPillars~\cite{lang2019pointpillars}? Range Sparse Net~\cite{sun2021rsn}?
It is easy to get overwhelmed by the diverse set of concepts present in these names as well as the variety in the architectures themselves. 

This level of variety at the macro-level is not observed in other areas, e.g., neural architectures developed for 2D images.
The root cause is the higher-dimensional nature of the data.
There are three major reasons in particular:
\begin{itemize}
    \item \emph{Views}: 2D images are captured by an egocentric photographer. A similar view exists for 3D, that is the perspective view, or range images. But when the scan is not egocentric, we have an unordered point set that can no longer be indexed by pixel coordinates. In addition, gravity makes the $z$ axis special, and often times a natural choice is to view an object from top-down. Each view has its unique properties and (dis)advantages. 
    \item \emph{Sparsity}: Images are dense in the sense that each pixel has an RGB value between 0 and 255. But in 3D, range images may have pixels that correspond to infinite depth. Also, objects typically occupy a small percentage of the space, meaning that when a scene is voxelized, the number of non-empty voxels is typically small compared with the total number of voxels.
    \item \emph{Neural operations}: Due to views and sparsity, 2D convolution does not always apply, resulting in more diverse neural operations.
\end{itemize}

Our first contribution in this paper is a \emph{unified framework} that can interpret and organize the variety of neural architecture designs, while adhering to the principles listed above.
This framework allows us to put existing designs in perspective and enables us to explore new designs.
The key idea is to factorize the entire neural network into a series of \emph{transforms} and \emph{layers}.
The framework supports four views (point, voxel, pillar, perspective) and two formats (dense, sparse), as well as the \emph{transforms} between them.
It is also possible to merge features from different views, building parallelism into the sequential stages.
But once a view-format combination is set, it restricts the types of \emph{layers} that can be applied. 
When visualized, this framework is a trellis, and any neural architecture corresponds to a connected subset of this trellis. 
We provide several examples of how popular architectures can be refactored and reproduced under this framework, proving its generality.

A direct benefit of this framework is that it can easily materialize into a search space, which immediately unlocks and enables NAS.
NAS stands for neural architecture search~\cite{zoph2016neural}, which tries to replace human labor and manual designs with machine computation and automatic discoveries.
Despite its success on 2D architectures~\cite{tan2019efficientnet}, its usage on 3D has been limited.
In this paper we conduct a principled NAS-for-3D explorations, by not only considering the micro-level (such as the number of channels), but also embracing the macro-level (such as transforms between various views and formats). 

We conduct our LidarNAS experiments on the 3D object detection task on the Waymo Open Dataset~\cite{sun2020scalability}.
Using regularized evolution~\cite{real2019regularized}, our search finds LidarNASNet, which outperforms the state-of-the-art RSN model~\cite{sun2021rsn} on both the vehicle and the pedestrian classes.
In addition to the superior accuracy and the competitive latency, there are also interesting observations about the LidarNASNet architecture itself.
First of all, though the search / evolution was conducted separately on vehicle and pedestrian, the found architectures have essentially the same high-level design concept.
Second, the modifications discovered by NAS coincidentally reflects ideas from human designs. 
We also analyze the hundreds of architectures sampled in the process and draw useful lessons that should inform future designs.

To summarize, the main contributions of this paper are:
\begin{itemize}
    \item A unified framework general enough to include a wide range of backbones for 3D data processing
    \item A search space and an algorithm challenging enough to cover both the micro-level and the macro-level
    \item A successful NAS experiment which leads to state-of-the-art performance on the Waymo Open Dataset
\end{itemize}

\section{Related Work}

\subsection{Neural Architectures for 3D}

We partition neural architectures for 3D into four categories, according to the primary view(s) used.
Since this paper studies \emph{backbone} design for 3D object detection, we will mostly cover detection but will also talk about segmentation and classification.

The first category is \textbf{top-down primary}, which includes voxel and pillar.
The main idea is to divide 3D points into 3D voxels~\cite{engelcke2017vote3deep,chen2017multi,yang2018pixor,zhou2018voxelnet,yan2018second,deng2020voxel} or 2D pillars~\cite{lang2019pointpillars}, which then become regular.
The advantage is that voxelization enables locality, which in turn enables convolution operations.
But the main limitation is memory consumption, which grows cubically (or quadratically).
This either limits the maximum detection range or sacrifices the voxelization granularity. 
Even if sparse operations may be used, for egocentric scans, the point densities at long-range and short-range are different, posing challenges in learning.

The second category is \textbf{point primary}, which treats the point cloud as unorganized sets.
Originally developed for classification and segmentation~\cite{qi2017pointnet,qi2017pointnet++}, the idea can also be used on detection~\cite{qi2019deep,ngiam2019starnet}.
The advantage is that it is more memory-friendly than voxelization based approaches.
However, its limitation is that the neural layers do not perform as well, possibly due to irregular coordinates.
In addition, to achieve locality, nearest neighbor search is typically needed for the input, which can be expensive.

The third category is \textbf{perspective primary}, operating directly on the range image~\cite{meyer2019lasernet,bewley2020range,chai2021point,fan2021rangedet}.
This is also very memory-friendly and can utilize powerful 2D convolution layers which have been extensively researched. 
However, as the depth can change drastically for adjacent pixels, these methods exhibit more difficulty in localizing the objects accurately, as well as handling occlusions.

The fourth and final category is \textbf{fusion} methods, which use two or more of the representations discussed above.
The fusion may be either sequential and parallel.
For example, RSN~\cite{sun2021rsn} sequentially performs foreground segmentation on the perspective view and delivers detection output on the top-down view.
PVCNN~\cite{liu2019point} and SPVCNN~\cite{tang2020searching} fuses information from the point view and the voxel view in a parallel fashion.
MVF~\cite{zhou2020end} fuses feature from perspective view, point view, and pillar view, also in a parallel fashion.
The hope is that fusion methods can combine the best of multiple worlds, which is why it is important to keep all options when doing architecture exploration.

\subsection{Neural Architecture Search}

Early works on neural architecture search primarily focused on the \textbf{search algorithm}. 
A variety of methods were introduced, including reinforcement learning~\cite{zoph2016neural,baker2016designing}, evolution~\cite{real2017large,real2019regularized}, performance prediction~\cite{liu2018progressive}, weight-sharing~\cite{pham2018efficient,liu2018darts}.
Essentially, different methods make different approximations about the search process.

These search algorithm explorations started on image classification.
The following phase consists of extending to other \textbf{tasks}, such as semantic segmentation~\cite{chen2018searching,liu2019auto} and object detection~\cite{xu2019auto,ghiasi2019fpn}.
For 3D tasks, NAS research has been done on medical imaging~\cite{zhu2019v,kim2019scalable,bae2019resource,wong2019segnas3d,yu2020c2fnas}.
However, the \emph{volumetric} CT scans are different from \emph{point} clouds, and as a result the search space is greatly simplified.
There are also works on 3D shape classification~\cite{ma2019auto,li2020sgas}, but their overall frameworks do not exceed that set by~\cite{liu2018darts}.
\cite{tang2020searching,li2020lc} is closer to our work, in the sense that it uses NAS to optimize for segmentation and detection on 3D scenes (KITTI~\cite{geiger2012we}).
But generalizing the terminology used in~\cite{liu2019auto}, we believe there is also a two-level hierarchy in 3D neural architecture designs, with the outer macro-level controlling the views of the data / features, and the inner micro-level being the specifics of the neural layers.
Under this terminology,~\cite{tang2020searching,li2020lc} keeps the macro-level fixed, while our search covers both.

\section{Unifying Neural Architectures for 3D}
\label{sec:unify}

\subsection{Philosophy}

In order to offer a unified interpretation of the growing variety of neural networks for 3D, we need to pinpoint their high-level design principles. 
Fortunately, we find these underlying principles to be surprisingly congruent, and we characterize them as: finding \emph{some neighborhood} of the 3D points and then \emph{aggregating information} within.
The ``aggregation'' part is typically done through some form of convolution and / or pooling. 
The ``neighborhood'' part has different choices:
\begin{itemize}
    \item PointNet~\cite{qi2017pointnet}: the neighborhood alternates between the point itself (MLP) and all points (max-pooling)
    \item PointNet++~\cite{qi2017pointnet++}: the neighborhood is an Euclidean ball with a certain radius
    \item VoxelNet~\cite{zhou2018voxelnet}: 3D neighborhood measured by Manhattan distance of Cartesian coordinates $(x, y, z)$
    \item PointPillars~\cite{lang2019pointpillars}: 2D neighborhood measured by Manhattan distance of (part of) Cartesian coordinates $(x, y)$
    \item LaserNet~\cite{meyer2019lasernet}: 2D neighborhood measured by Manhattan distance of pixel coordinates $(i, j)$
\end{itemize}

These common ``neighborhood'' choices have been typically expressed through the views of the data / features: point, voxel, pillar, perspective.
We point out that there have been and will be more views being proposed, which is why we feel the ``neighborhood'' interpretation is more generic.
Notably, different data views can \emph{transform} between each other back and forth. 
However, once the data view is determined, it \emph{restricts} the type of \emph{layers} that can be applied. 
This factorization of ``transforms'' and ``layers'' as well as their relationship will be reflected in our framework described next.

\subsection{A Unified Framework}

In this subsection, we build upon the aforementioned high-level ideas and describe the main framework we use to think about neural architectures throughout this work. 
We describe its different levels of detail from fine to coarse.

\paragraph{Views and formats}

We consider a total of four views (point, pillar, voxel, perspective) and up to two data formats (dense and sparse):
\begin{itemize}
    \item \textbf{Point}: The features for all $N$ 3D points are stored in a matrix of size $[N, C]$, where $C$ is the number of channels. The Cartesian coordinates $(x, y, z)$ for each point are stored in a separate matrix of size $[N, 3]$, where the indices of the points are aligned between the two matrices.
    \item \textbf{Pillar}: In this view, we store a fixed-length feature for each pillar when viewing the scene from top-down. We allow the pillar view to be either dense or sparse. If dense, the features are stored in a tensor of size $[B, X, Y, C]$, where $B$ is the batch size, $X$ and $Y$ are the number of pillars along the corresponding dimension. If sparse, the features are stored in a matrix of size $[N, C]$, where $N$ is the number of \emph{non-empty} pillars and a separate matrix of size $[N, 3]$ is used to store the indices (both batch and spatial) of these non-empty pillars. In both data formats, unlike the point view, the Cartesian coordinates of each pillar('s center) can be easily calculated from its spatial index (\texttt{origin + index * pillar size}).
    \item \textbf{Voxel}: Different from the pillar view, the voxel view partitions the scene along all three spatial dimensions. The additional partition along the $z$ axis makes fitting a tensor of size $[B, X, Y, Z, C]$ into memory very challenging. Therefore, in this work we only consider the sparse format for the voxel view. Features are stored in a matrix of size $[N, C]$, where $N$ is the number of \emph{non-empty} voxels. A separate matrix of size $[N, 4]$ is used to store their indices (both batch and spatial). 
    \item \textbf{Perspective}: For egocentric 3D scans or RGBD images, simply using the original perspective view is a natural choice. We consider both the dense and sparse formats for this view. If dense, features are stored in a tensor of size $[B, H, W, C]$, where $H$ and $W$ consist of the size of the range image. A separate tensor of size $[B, H, W, 3]$ is used to store the Cartesian coordinates of each pixel on the range image. If sparse, features are stored in a matrix of size $[N, C]$ and Cartesian coordinates a separate matrix of size $[N, 3]$, similar to the point view.
\end{itemize}

\paragraph{Transforms}

Now that the views and formats are established, the framework shall be general enough to include possible transforms from one to the other.
The transforms are intended to be lightweight: powerful neural feature update is a non-goal. 
Since there are totally six possible representations (four views, with pillar and perspective having two formats), we have up to $6^2 = 36$ different transforms. 
Though this number may seem daunting, some of these transforms have more familiar and friendly names.
For example, the transform to itself is identity.
The one from a sparse format to its dense counterpart is densification (by padding zero vectors). 
The point to voxel transform is voxelization. 
The reverse transform is devoxelization. 
The point to perspective transform is projection. 

\begin{figure}[t]
    \centering
    \includegraphics[width=\textwidth]{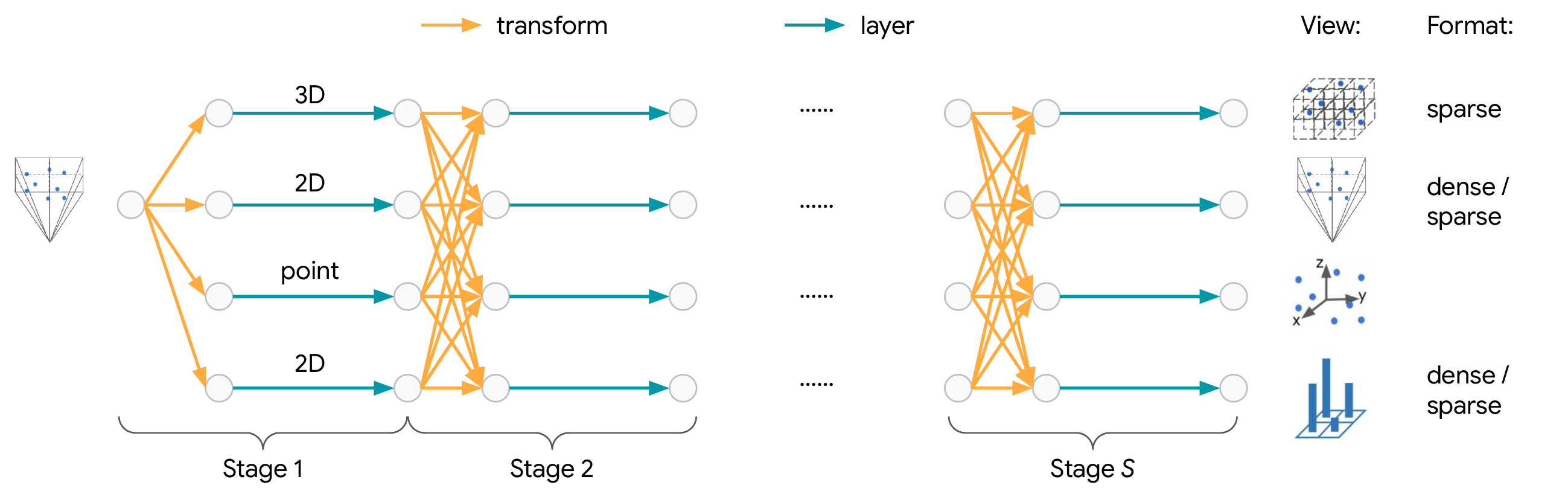}
    \caption{The LidarNAS framework for interpreting neural architectures on 3D point clouds. The entire backbone consists of $S$ stages. Each stage consists of view \& format transforms followed by corresponding neural layers. Within this framework, a backbone architecture corresponds to a \emph{connected subset} of the $S$-stage trellis. }
    \label{fig:framework}
\end{figure}

\paragraph{Layers}

Once a view-format combination is set, we can apply neural layers to update the features. 
Generally, we do not put constraint on the number or form of the layers: it can be as simple as a one-layer convolution, or as complicated as an entire U-Net~\cite{ronneberger2015u}. 
But the one constraint is that it conforms to the view-format combination for both its input and output.
This is because, for instance, 2D convolution cannot be applied on 3D inputs; sparse convolution does not work on dense features. 
Notably, 2D layer implementations can interchangeably work for both the pillar view and the perspective view.

\paragraph{Stages}

Putting these concepts together, we define a stage to be the sequential pair of possible transforms and their associated layers. 
\cref{fig:framework} visualizes the concatenation of $S$ stages.
Within this framework, the backbone of a neural network for 3D corresponds to a \emph{connected subset of this $S$-stage trellis}.
A head can then be added to the end to perform 3D classification / detection / segmentation.

We emphasize that the word choice is ``subset'' but not ``path'', meaning that a stage can have more than one view present.
This makes our framework more general, as it supports not only sequential designs but also parallel ones.
Consequently, we may have multiple different views in stage $s - 1$ transforming to the same view in stage $s$.
In these cases, after applying individual transforms, we merge these transformed features through either concatenation (default in this work) or summation. 

\subsection{Inclusion of Existing Designs}

\begin{figure}[t]
    \centering
    \begin{subfigure}[b]{0.49\linewidth}
    \includegraphics[width=\linewidth]{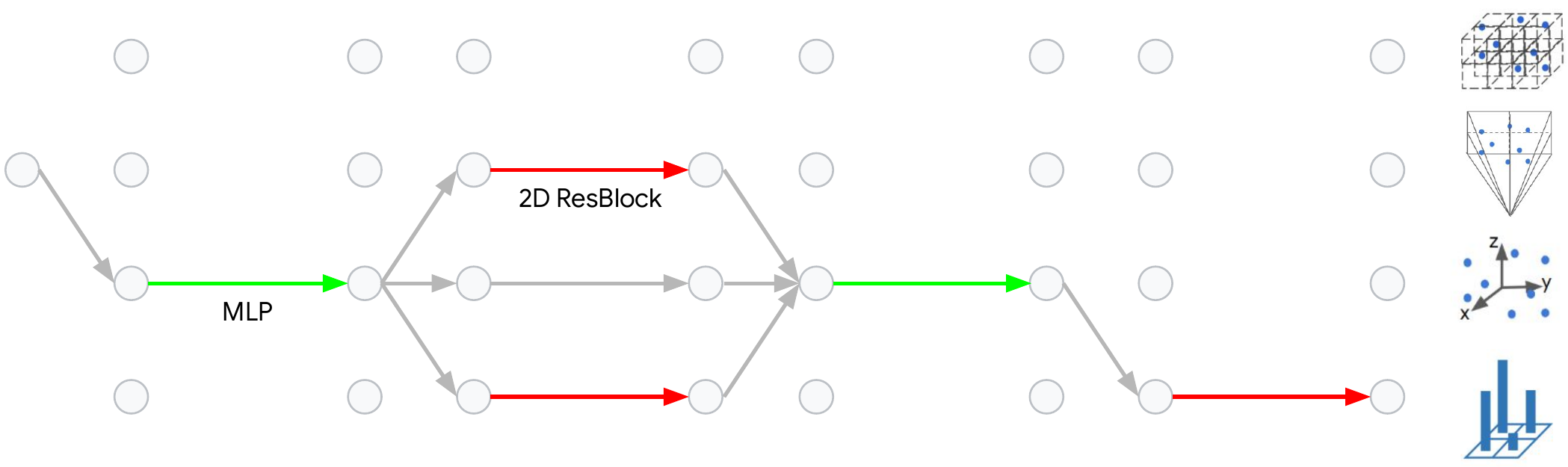}
    \caption{Multi-View Fusion~\cite{zhou2020end}}
    \end{subfigure}
    \begin{subfigure}[b]{0.49\linewidth}
    \includegraphics[width=\linewidth]{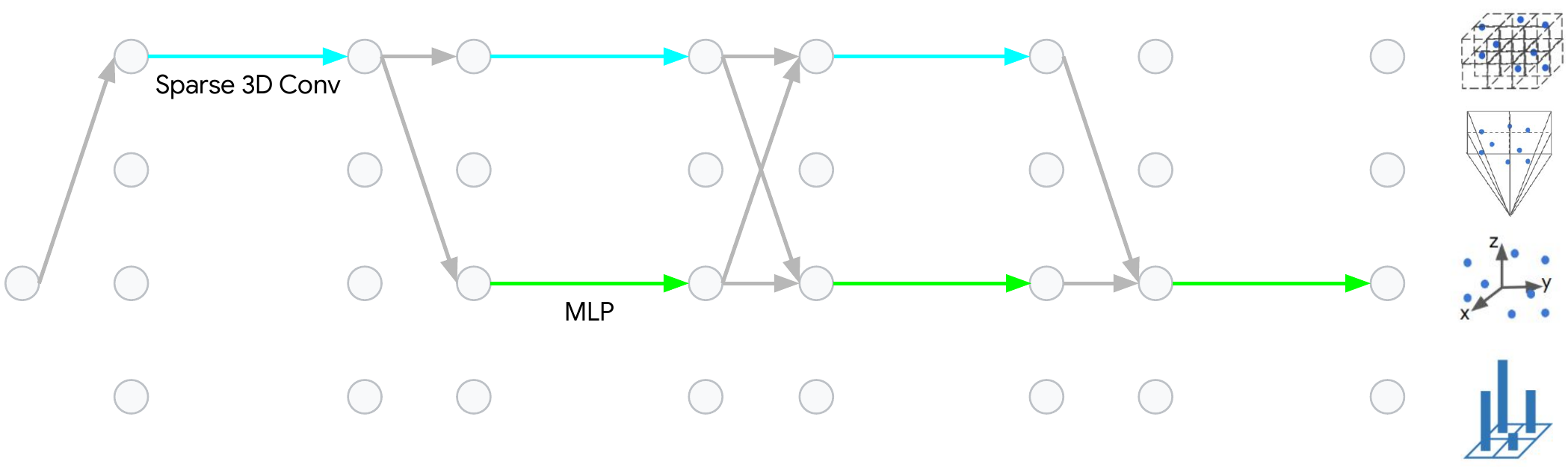}
    \caption{Sparse Point-Voxel~\cite{tang2020searching}}
    \end{subfigure}
    \caption{Examples of how existing designs may be interpreted within the LidarNAS framework.}
    \label{fig:existing}
\end{figure}

In \cref{fig:existing} and \cref{fig:lidarnasnet}, we visualize several examples of how existing designs may be interpreted within the framework described above. 
Our framework is flexible enough to cover both entirely sequential designs such as Range Sparse Net~\cite{sun2021rsn} and more parallel designs such as Multi-View Fusion~\cite{zhou2020end}.
In addition to these networks developed for 3D detection, it can also explain those beyond, such as SPV~\cite{tang2020searching}.
More architecture designs fit in, including but not limited to~\cite{zhou2018voxelnet,yan2018second,lang2019pointpillars,ngiam2019starnet,wang2020pillar}, but we skip visualization due to space limitations.

\section{Searching Neural Architectures for 3D}
\label{sec:search}

The framework described in the previous section brings many benefits, one of which is the potential to search novel and better architectures.
This section focuses on how the framework materializes into a search space (\cref{sec:sspace}), as well as our choice of search algorithm (\cref{sec:evolution}).

\subsection{From Framework to Search Space}
\label{sec:sspace}

\cref{fig:existing} demonstrated how, at a high level, various architectures fall within the LidarNAS framework.
But delving into the details, specific implementations of the modules are going to differ across works.
While this is very much expected and understandable, the variety and freedom in ``layers'' alone would make constructing a meaningful search space infeasible.
We now discuss how we materialize the framework into a search space by making specific choices.

\paragraph{Transforms}

Among the $36$ possible transforms, we did not implement the transforms from pillar to voxel, as nothing more can be done other than copying the same features along the $z$ axis.
From the voxel view, our implementation only supported transforms to the pillar view.
Supporting $31 / 36$ transforms is still high coverage.

\paragraph{Layers}

We need at least one type of neural layer for each of the following: point, 2D dense, 2D sparse, 3D sparse.
Our search space picked one representative for each:
\begin{itemize}
    \item \textbf{Point}: Multiple layers of dense-normalization-ReLU. The normalization can either be batch normalization~\cite{ioffe2015batch} or layer normalization~\cite{ba2016layer}. The number of units $F$ in the dense layers is a hyperparameter that can be searched.
    \item \textbf{2D dense}: A U-Net~\cite{ronneberger2015u} with residual blocks~\cite{he2016deep}. We use up to five downsampling and upsampling scales. The number of channels for each scale are $[F, 4F, 8F, 8F, 16F]$ with $F$ being the hyperparameter that can be searched. The number of blocks per scale is $2$ except for the highest resolution scale which is $1$.
    \item \textbf{2D sparse}: Also a U-Net with residual blocks, except that each convolution is a sparse convolution (kernel size $3 \times 3$). We use up to $3$ downsampling and upsampling scales, and the number of blocks are $[1, 2, 3]$ and $[0, 2, 2]$. We use the same number of channels $F$ for all downsampling and upsampling blocks.
    \item \textbf{3D sparse}: Also a U-Net with residual blocks, except that 3D sparse convolution is used. The kernel size can either be $3 \times 3 \times 3$ or $3 \times 3 \times 1$, and the corresponding stride for each scale is $2 \times 2 \times 2$ or $2 \times 2 \times 1$. The other details follow the 2D sparse case above.
\end{itemize}

These choices of layer specifics, especially those for 2D and 3D, try to exactly follow RSN~\cite{sun2021rsn}.

\paragraph{Stages}

Our search space considers $S = 3$ stages.
For simplicity, we also have the constraint that the last stage can only have one view.
Inspired by RSN, we add the option to perform foreground segmentation immediately after the first perspective branch that appears.

\subsection{Regularized Evolution}
\label{sec:evolution}

We choose regularized evolution to be our search algorithm, which follows~\cite{real2019regularized}.
Compared against other major classes of NAS methods, evolution arguably makes the least amount of approximations, which is desirable especially since we are exploring a less explored task and a complicated search space.
We do not use weight-sharing NAS for GPU memory considerations.
3D tasks are understandably more memory intensive than 2D tasks, and the batch size on each GPU was already small ($< 10$). 
However, even the best weight-sharing NAS (a recent example is~\cite{bender2020can}) implementations require $2 - 3 \times$ extra GPU memory. 

Our mutation algorithm works by first randomly selecting a stage $s$ and then randomly applying one of the following six mutation choices to this stage:
\begin{itemize}
    \item \textit{Add a view}: if the stage does not have all four views, then randomly add a view not yet present in this stage. A random view present in the previous stage is selected as its predecessor. A random view present in the next stage is selected as its successor. A default layer of the corresponding type is used for this addition. The number of channels for all layers in this stage are halved.
    \item \textit{Remove a view}: if the stage has more than one view, then randomly remove an existing view. Usage of the removed view in the next stage is also removed. The number of channels for all layers in this stage are doubled.
    \item \textit{Switch the view}: if the stage has exactly one view, then switch the view to another. All usage of the old view in the next stage is changed to the new view. 
    \item \textit{Adjust the pillar / voxel size}: a key parameter in many of the transforms is the pillar / voxel size. Multiply the pillar / voxel size by either $0.8$ or $1.2$ for all views.
    \item \textit{Adjust the number of channels}: multiply the number of channels for all layers in the stage by either $0.8$ or $1.2$.
    \item \textit{Adjust the layer progression}: 
    \begin{itemize}
        \item Point: Either increase or decrease the number of dense-normalization-ReLU by $1$.
        \item 2D dense: Either increase or decrease the number of scales by $1$.
        \item 2D / 3D sparse: Increase or decrease the number of downsampling / upsampling scales by $1$.
    \end{itemize}
\end{itemize}
If a mutation fails (e.g., if the precondition does not hold, such as trying to remove a view when the stage only has one view), the algorithm mutates again until it succeeds.

The first four mutation choices focus on the ``transform'' aspect of a stage, while the last two mutation choices focus on the ``layer'' aspect.
This level of coverage and variety makes the search comprehensive yet challenging.

\section{Experimental Results}

\subsection{Experimental Setting}

We perform 3D object detection experiments on the challenging Waymo Open Dataset~\cite{sun2020scalability}.
It provides LiDAR scans in the range image form, which makes experiments on the perspective view much more natural and convenient.
It contains $1150$ LiDAR sequences with $798$ train, $202$ validation, and $150$ test ones.
Each sequence is $20$ seconds at $10$ frames per second.
Experiments are conducted on both the vehicle and the pedestrian classes, using the official evaluation metrics of 3D / BEV AP.

\subsection{Existing Architectures under LidarNAS}
\label{sec:existing}

In this subsection, we use the LidarNAS framework (\cref{sec:unify}) to reimplement several existing neural architectures for 3D. 
The goal here is to prove the generality and correctness of the LidarNAS framework interpretation, as well as validate our implementation of individual modules.

We selected four existing architectures: RSN~\cite{sun2021rsn}, PointPillars~\cite{lang2019pointpillars}, LaserNet~\cite{meyer2019lasernet}, and MVF++~\cite{qi2021offboard}. 
These are selected to cover a variety of views as well as topology. 
Note that the LidarNAS framework focuses on \emph{backbone} design.
In our reimplementation, we use an anchor-free detection \emph{head} that is the same as RSN if the backbone output is sparse voxels but also works for pillar and perspective views (details are described in the supplementary material). 
This means that our RSN reimplementation is exact but the others are not, and we add the suffix ``-like'' to indicate this difference.

\begin{table}[t]
    \centering
    \scalebox{0.78}{
    \begin{tabular}{ccccccccc | c}
        \toprule
        model & class & frame & device & batch & steps & lr & voxelization (m) & LidarNAS AP & previous AP \\
        \midrule
        \multirow{2}{*}{RSN-exact} & Veh & $3$ & GPU & $2 \times 16$ & $120$k & $0.006$ & $0.2 \times 0.2 \times 0.2$ & $77.2$ & $77.2$~\cite{sun2021rsn} \\
         & Ped & $3$ & GPU & $3 \times 16$ & $120$k & $0.006$ & $0.1 \times 0.1 \times 20.0$ & $79.1$ & $79.1$~\cite{sun2021rsn} \\
        \midrule
        \multirow{2}{*}{PointPillars-like} & Veh & $1$ & GPU & $2 \times 16$ & $120$k & $0.006$ & $0.32 \times 0.32$ & \colorbox{SpringGreen}{$69.3$} & $63.3$~\cite{sun2021rsn} / $60.3$~\cite{qi2021offboard} \\
        & Ped & $1$ & GPU & $3 \times 16$ & $120$k & $0.006$ & $0.32 \times 0.32$ & \colorbox{Goldenrod}{$66.1$} & $68.9$~\cite{sun2021rsn} / $60.1$~\cite{qi2021offboard} \\
        \midrule 
        \multirow{2}{*}{LaserNet-like} & Veh & $1$ & GPU & $1 \times 16$ & $360$k & $0.001$ & - & \colorbox{Goldenrod}{$47.1$} & $52.1$~\cite{sun2021rsn} / $56.1$~\cite{chai2021point} \\
        & Ped & $1$ & GPU & $1 \times 16$ & $240$k & $0.003$ & - & \colorbox{Goldenrod}{$59.0$} & $63.4$~\cite{sun2021rsn} / $62.9$~\cite{chai2021point} \\
        \midrule 
        \multirow{2}{*}{MVF++-like} & Veh & $1$ & TPU & $2 \times 128$ & $43$k & $0.003$ & $0.32 \times 0.32$ & $73.6$ & $74.6$~\cite{qi2021offboard} \\
        & Ped & $1$ & TPU & $2 \times 128$ & $43$k & $0.003$ & $0.32 \times 0.32$ & \colorbox{Apricot}{$70.4$} & $78.0$~\cite{qi2021offboard} \\
         \bottomrule
    \end{tabular}
    }
    \caption{A diverse set of existing 3D detection architectures under the LidarNAS framework. The second number in the batch size multiplication is the number of GPUs / TPU shards. The metric (last two columns) is L1 3D AP.}
    \label{tab:reproduce}
\end{table}

\cref{tab:reproduce} summarizes the results, using L1 3D AP. 
Key hyperparameter values are also provided.
We use no color if our reimplementation is within $1\%$ absolute of the previously reported number; green if higher than $>5\%$; yellow if lower than $\leq 5\%$; and red if lower than $>5\%$.
Considering the diversity of these architectures, overall we consider our reimplementation to be acceptable and successful, validating our implementation of (some of the) transforms and layers modules.
Notice that our implementation can support multi-frame, as well as both GPU and TPU.

Looking into individual neural architectures, our reproduction of RSN is exact.
Interestingly, PointPillars-like significantly outperforms previous reports on the vehicle class. 
This is an important reminder that revisiting previous architectures may be necessary and beneficial, as they may still be competitive when coupled with latest developments in other areas (e.g., anchor-free detection head). 
However, the performance on the pedestrian class is slightly worse.
This is also observed on MVF++-like, where the vehicle class is within $1\%$ but the pedestrian class is significantly worse. 
Our hypothesis is that comparatively speaking, our detection head is better suited on larger objects but struggles more on smaller objects.
Finally, our LaserNet-like performs noticeably worse than any network that detects on the top-down view (meaning pillar or voxel), despite training for $2-3 \times$ longer steps.
This proves that detection from the perspective view needs more specialized operations, such as those described in the original paper, or some recent developments~\cite{chai2021point,fan2021rangedet}.

\subsection{Searching for New Architectures}

In this subsection, we perform and analyze neural architecture search experiments, using the search space and algorithm described in \cref{sec:search}.

\begin{table}[t]
    \centering
    \scalebox{0.84}{
    \begin{tabular}{ccc | ccc | ccc}
        \toprule
        & & & \multicolumn{3}{c|}{Vehicle} & \multicolumn{3}{c}{Pedestrian} \\
        model & year & frame & 3D AP & BEV AP & latency & 3D AP & BEV AP & latency \\
        \midrule
        LaserNet~\cite{meyer2019lasernet} & CVPR 19 & & $52.1$ & $71.2$ & $64.3$ & $63.4$ & $70.0$ & $64.3$ \\
        PointPillars~\cite{lang2019pointpillars} & CVPR 19 & & $63.3$ & $82.5$ & $49.0$ & $68.9$ & $76.0$ & $49.0$ \\
        PV-RCNN~\cite{shi2020pv} & CVPR 20 & & $70.3$ & $83.0$ & - & - & - & - \\
        Pillar-based~\cite{wang2020pillar} & ECCV 20 & $1$ & $69.8$ & $87.1$ & $66.7$ & $72.5$ & $78.5$ & $66.7$ \\
        \gray{PV-RCNN~\cite{shi2020pvwod}} & \gray{WOD 20} & \gray{$2$} & \gray{$77.5$} & \gray{-} & \gray{$300$} & \gray{$78.9$} & \gray{-} & \gray{$300$} \\
        RCD~\cite{bewley2020range} & CoRL 20 & $1$ & $69.0$ & $82.1$ & - & - & - & - \\
        MVF++~\cite{qi2021offboard} & CVPR 21 & $1$ & $74.6$ & $87.6$ & - & $78.0$ & $83.3$ & - \\
        \gray{CenterPoint~\cite{yin2021center}} & \gray{CVPR 21} & \gray{$2$} & \gray{$76.7$} & \gray{-} & \gray{-} & \gray{$79.0$} & \gray{-} & \gray{-} \\
        PPC~\cite{chai2021point} & CVPR 21 & & $65.2$ & $80.8$ & - & $75.5$ & $82.2$ & - \\
        RangeDet~\cite{fan2021rangedet} & ICCV 21 & $1$ & $72.9$ & - & - & $75.9$ & - & - \\
        \midrule
        PointPillars-like$^\mathsection$ & & 1 & 67.6 & 85.3 & - & - & - & - \\
        LidarNASNet-P (ours) & & 1 & \bf{73.2} & \bf{88.2} & - & - & - & - \\
        \midrule
        RSN~\cite{sun2021rsn} & CVPR 21 & $1$ & $75.2$ & $87.7$ & $46.5^\dagger$ & $77.1$ & $81.7$ & $21.0^\dagger$ \\
        LidarNASNet-R (ours) & & $1$ & $\bf{75.6}$ & $\bf{88.6}$ & $49.3^\dagger$ & $\bf{77.4}$ & $\bf{82.0}$ & $22.6^\dagger$ \\
        \bottomrule
    \end{tabular}
    }
    \caption{3D object detection results for the vehicle and pedestrian classes on the Waymo Open Dataset validation set. The AP is difficulty L1. The unit of latency is ms. Multi-frame models are \gray{grayed}. $^\mathsection$: Slightly different from PointPillars-like in \cref{tab:reproduce}, because here we have to swap the original layers with the U-Net explained in \cref{sec:sspace}. $^\dagger$: our measurement using identical setting: average on $10$ scenes, each has more than $100$ vehicles / pedestrians.}
    \label{tab:leaderboard}
\end{table}

\subsubsection{Evolving past the state-of-the-art}

Based on the analysis above, picking a random architecture as the starting point would take much longer time for the performance to ramp up, so we use \emph{warm starting}~\cite{so2019evolved,so2021primer} to speed up and save up. 
Each search lasts $100$ architectures, each trained using batch size $2 \times 8$ GPUs for $12$k steps ($10\%$ of the standard number of steps) using cosine learning rate. 
All architectures operate on single-frame. 
The population size and tournament size for the regularized evolution algorithm are $20$ and $5$ respectively. 
We also measure the V100 latency of the network on a (random) training batch immediately after $11$k training steps.
The measurement is taken close to the end of the training because for architectures that perform foreground segmentation, the latency may change throughout training. 
We comment that this search phase latency measurement is noisy, not only because the data batch is random, but also because the scheduler may allocate a GPU shared with other jobs.
Regardless, we use \texttt{100 * L1 3D AP - 0.5 * latency in ms}\footnote{We empirically picked these multipliers; did not tune them heavily.} as the objective to guide the evolution.
Once an architecture is identified, we increase the per-GPU batch size from $2$ to $5$, and train for $120$k steps as the final evaluation.

We conduct a separate search / evolution from three different starting points: PointPillars-like vehicle, RSN CarXL, and RSN PedL\footnote{We skipped PointPillars-like pedestrian, because the corresponding number in \cref{tab:reproduce} is yellow not green.}.
We name our found architecture LidarNASNet-P / R depending on whether the starting point was PointPillars-like or RSN, and compare them against other models in \cref{tab:leaderboard}.

We first compare LidarNASNet-P against PointPillars-like, the evolution baseline.
The L1 3D AP improves from $67.6$ to $73.2$, with a significant gap of $+5.6$.
The gain on BEV AP is also significant at $+2.9$.
The large improvement and competitive end result clearly showcase the effectiveness of our search.
When running the same evolution from RSN, LidarNASNet-R outperforms by $0.4$ and $0.3$ 3D AP on vehicle and pedestrian respectively, and the gains on BEV AP are even larger.
As we will see soon, LidarNASNet-R has an additional branch, so the latency is higher, but only slightly. 
To put this in perspective, we did an ablation study\footnote{In fact this architecture was sampled / discovered during our evolution.} where we increase the number of channels in the sparse U-Net of RSN for vehicle (from $64$ to $91$) to reach AP parity with LidarNASNet-R.
The latency of this architecture is $60.8$ms, which is significantly higher.

Comparing against other architectures, LidarNASNet-R also performs very competitively.
Not only is this reflected in the superior AP especially among single-frame models, but also in the small latency.
We reiterate that our total search cost is about $80$ GPU days, which is only $10$ times the cost of training a single RSN ($8$ GPU days).

\subsubsection{Visualizing and analyzing LidarNASNet}

\begin{figure}[t]
    \centering
    \begin{subfigure}[b]{0.49\linewidth}
    \includegraphics[width=\linewidth]{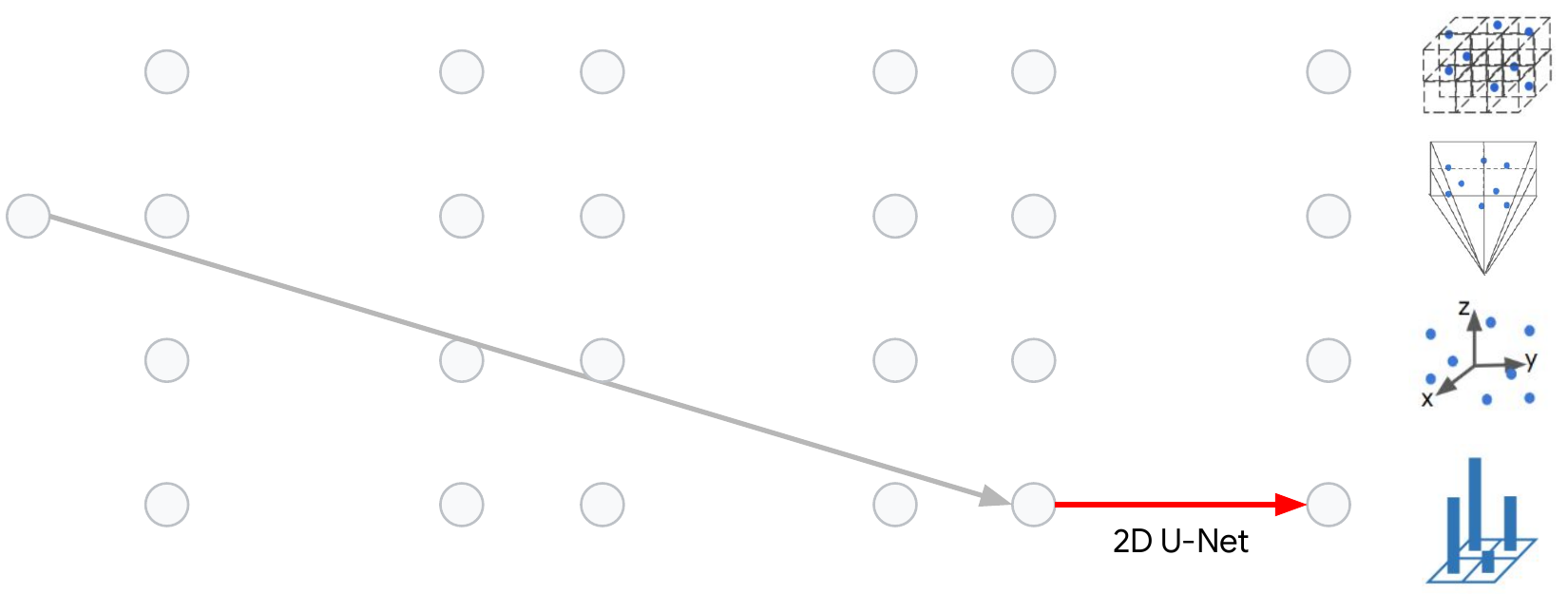}
    \caption{PointPillars-like}
    \end{subfigure}
    \begin{subfigure}[b]{0.49\linewidth}
    \includegraphics[width=\linewidth]{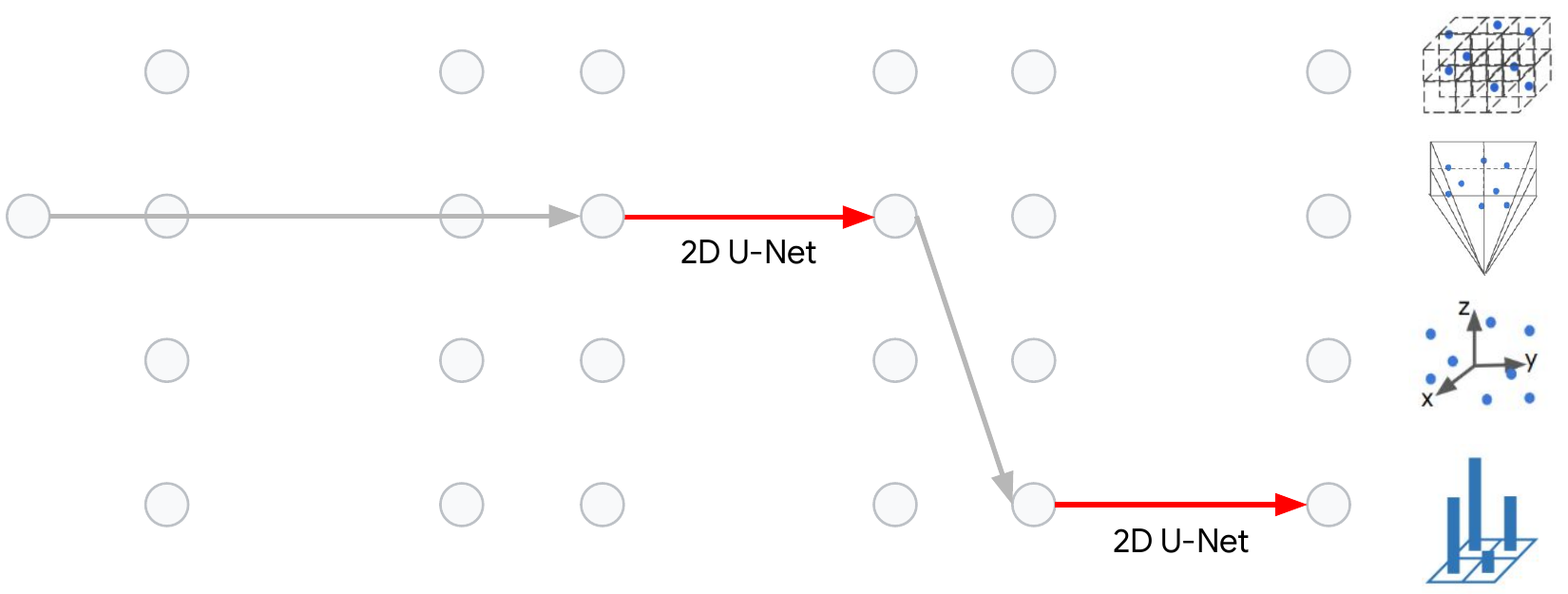}
    \caption{LidarNASNet-P}
    \end{subfigure}
    \begin{subfigure}[b]{0.49\linewidth}
    \includegraphics[width=\linewidth]{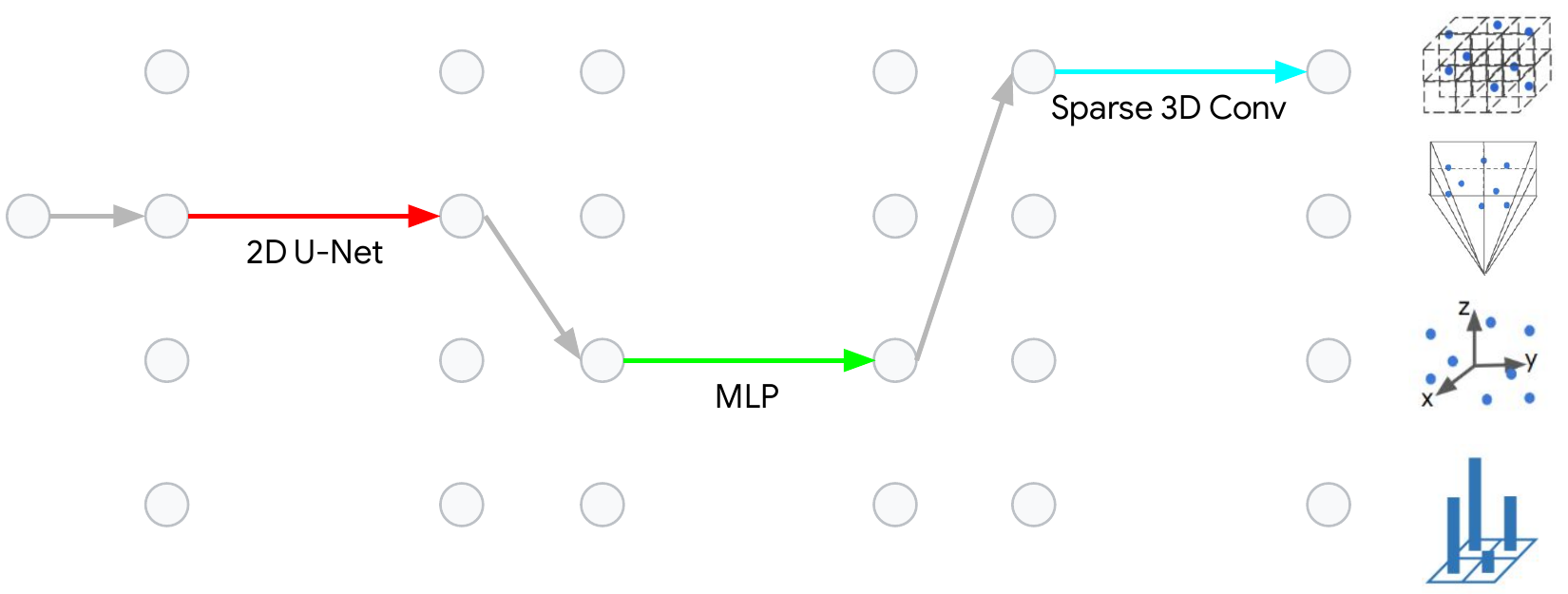}
    \caption{Range Sparse Net~\cite{sun2021rsn}}
    \end{subfigure}
    \begin{subfigure}[b]{0.49\linewidth}
    \includegraphics[width=\linewidth]{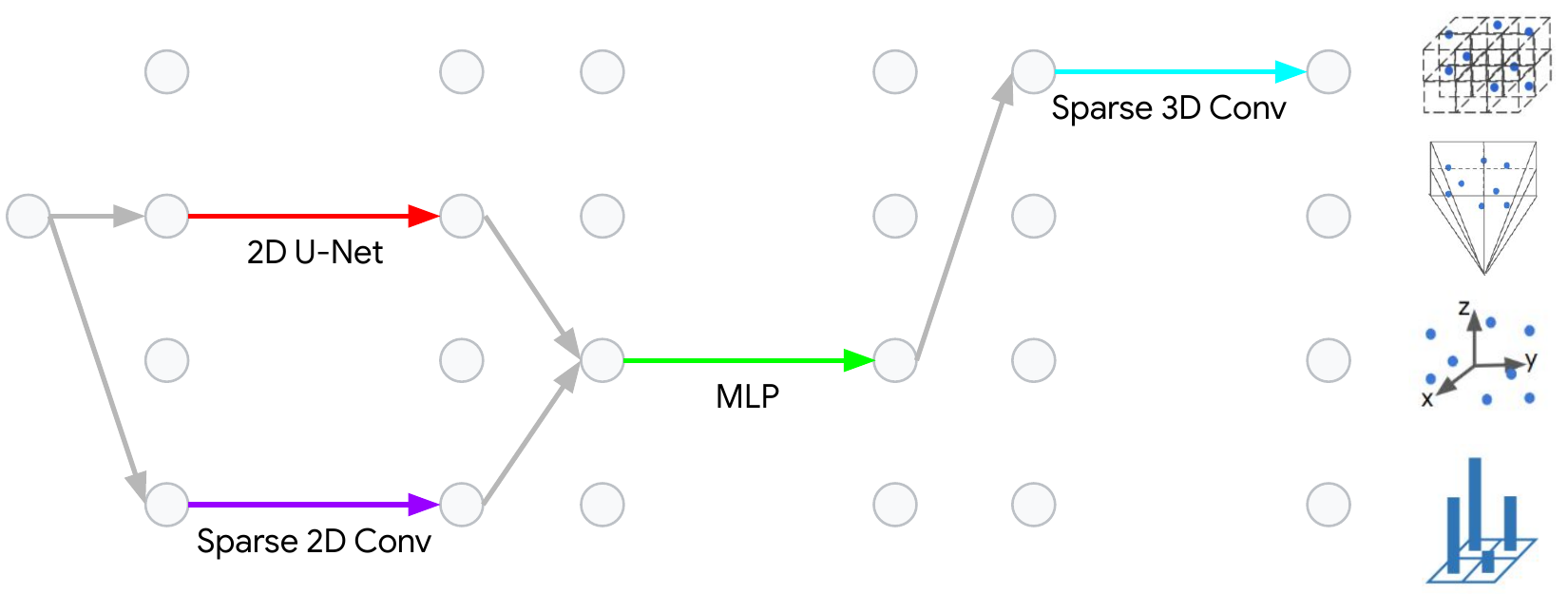}
    \caption{LidarNASNet-R}
    \end{subfigure}
    \caption{The \emph{macro-level} architecture of the found LidarNASNet-P / R. Note that the illustration in (d) applies for \emph{both} vehicle and pedestrian. It adds a sparse convolution on the pillar view in the first stage, utilizing all four views and two formats considered in this work.}
    \label{fig:lidarnasnet}
\end{figure}

We visualize the macro-level architecture of LidarNASNet in \cref{fig:lidarnasnet}.
We start by discussing LidarNASNet-P.
At the macro-level, the evolution decided to add a 2D U-Net that enhances the features for each range image pixel before voxelization to the pillar view.
This change alone improves the 3D AP from $67.6$ to $72.3$.
The evolution also learned to increase the voxelization granularity from $0.32 \times 0.32$ to $0.25 \times 0.25$, which is a micro-level change that is not reflected in \cref{fig:lidarnasnet}.
This change further improves the 3D AP to the $73.2$ reported in \cref{tab:leaderboard}.

For LidarNASNet-R, notice that though the search was conducted separately for the vehicle and pedestrian class, \emph{the same macro-level architecture design was found}, which is a positive signal regarding the generality of the found design.
Specifically, LidarNASNet-R adds a pillar view in the first stage, as well as the associated sparse 2D U-Net.
The idea of adding a pillar view resembles MVF~\cite{zhou2020end,qi2021offboard} (though the sparse format is used here while MVF did not consider sparse operations), making LidarNASNet-R a hybrid between RSN and MVF, two very successful human designs.
The voxelization granularity of this sparse 2D U-Net is $0.32 \times 0.32$, and the number of channels $F$ is $16$.
In the vehicle variant, the number of channels for the original perspective view is halved (from $16$ to $8$).
In the pedestrian variant, this is reduced even more aggressively (from $16$ to $3$).

\subsubsection{The search space is challenging}

\begin{wrapfigure}{r}{0.4\linewidth}
    \centering
    \includegraphics[width=0.75\linewidth]{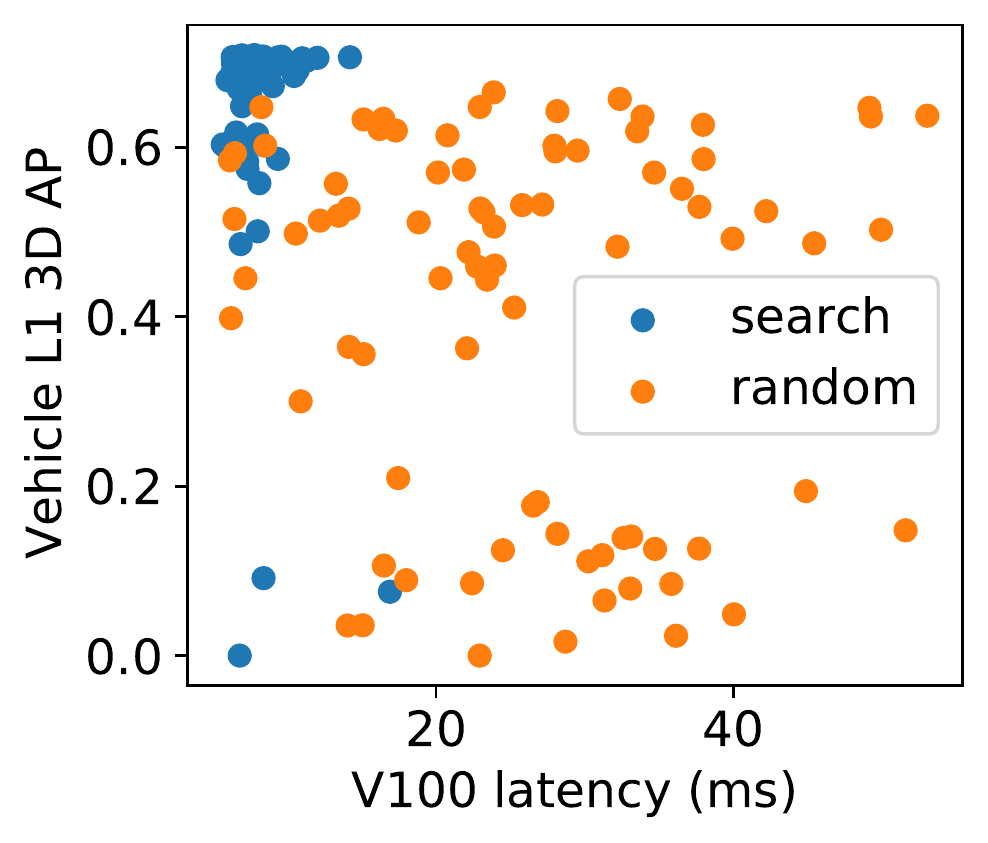}
    \caption{Randomly sampled architectures (orange) in the LidarNAS search space have worse average and higher variance, calling for warm starting (blue).}
    \label{fig:search_vs_random}
\end{wrapfigure}

A common critique on some of the NAS literature is that the search space can be ``easy'' in the sense that even random sampling of architectures (and taking the argmax) can find high-quality architectures indistinguishable from those found by NAS~\cite{li2020random}. 
We prove our search space is not trivial, by training $100$ architectures randomly generated by the procedure detailed in the supplementary material.
Figure~\ref{fig:search_vs_random} shows the side-by-side comparison of these random architectures against our LidarNAS evolution.
It is clear that randomly sampled architectures have much worse qualities, in terms of both detection AP and latency. 
Not only does this illustrate that the search space we consider is challenging and nontrivial, but also justifies our use of warm starting.

\subsubsection{Lessons from the sampled architectures}

In addition to fixating on the top-performing architecture, there are also lessons to be learned in the hundreds of architectures sampled.
We now choose a few angles to analyze these data.

Using the LidarNAS evolution data points, we investigate architectures that only mutated the ``layer'' aspect (i.e. the last two mutation choices in \cref{sec:evolution}) versus the rest.
The AP standard deviation of the two subsets are $0.04$ and $0.14$ respectively, which confirms that on average, mutating ``transforms'' results in more aggressive changes than mutating ``layers'' only.

\begin{wrapfigure}{r}{0.4\linewidth}
    \centering
    \includegraphics[width=0.75\linewidth]{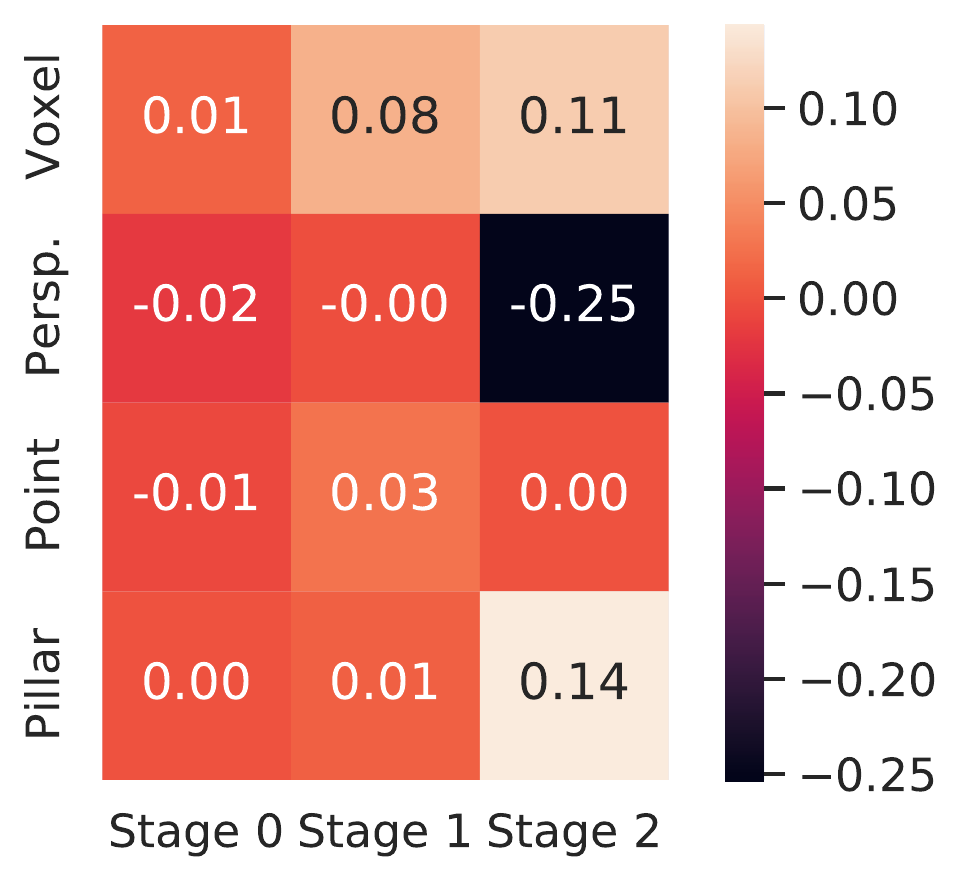}
    \caption{Linear regression from the presence of individual views and stages to detection AP.}
    \label{fig:linear_regression}
\end{wrapfigure}

Using the random architectures data points, we study which views and stages have the most direct effect on detection quality.
Specifically, we run a linear regression from the 12-dimensional binary feature indicating whether the corresponding branch exists in the architecture to the L1 3D AP. 
The coefficients are visualized in \cref{fig:linear_regression}.
By comparing the columns, it is clear that later stages have a much more direct influence on the detection AP than earlier stages.
By comparing the rows of the last column, top-down views (voxel and pillar) positively influence the detection AP, while the perspective view impacts it negatively.
This again resonates with the belief that detection from the perspective view tends to be more challenging and requires more specialized treatment.

Using the random architectures data points, we also study the effect of dense vs sparse on latency.
Recall that in our LidarNAS framework, the perspective view and the pillar view are the two that allow both dense and sparse formats.
For each view, we run a linear regression to latency from a 3-dimensional feature, indicating the total number of empty / dense / sparse branches.
For the perspective view, the coefficients are $[-5.24, -1.65, 6.90]$.
For the pillar view, the coefficients are $[-3.03, 3.06, -0.03]$.
The first coefficient is the most negative for both, which is expected, because the more empty branches you have, the smaller the latency is.
Interestingly, the coefficients reveal that on the perspective view, using more sparse branches results in larger latency, whereas on the pillar view, using more sparse branches results in smaller latency. 
This shows that sparse operations can offer speedup but not always: it depends on whether the view inherently has high sparsity.

\section{Conclusion}

This paper aims to achieve two goals for neural architecture research for 3D: first, a unified framework that summarizes and organizes existing designs, and second, an architecture search exploration enabled by this framework.
We demonstrate the generality of our LidarNAS framework, not only through pictorial illustration, but also through empirical experiments.
Then, we successfully and automatically discovered LidarNASNet, which achieves state-of-the-art results on the Waymo Open Dataset 3D object detection.
The searched architecture is interesting: not only is it identical when searching on two different classes, but also embodies and reaffirms shades from existing designs.

There are still many limitations in this work, and we look forward to addressing them in future research.
First, while the \emph{transforms} coverage is fairly complete, the \emph{layers} currently implemented do not capture much diversity, and we shall add more powerful layer choices into the search space, such as Transformers~\cite{zhao2021point,engel2021point,mao2021voxel}.
Second, all search experiments are single-frame; in extending to multi-frame, challenges include more memory pressure and the additional complication over which stage to perform temporal fusion.

% \clearpage
% ---- Bibliography ----
%
% BibTeX users should specify bibliography style 'splncs04'.
% References will then be sorted and formatted in the correct style.
%
\bibliographystyle{splncs04}
\bibliography{5318}

\clearpage
\appendix

\section{Anchor-Free Detection Head}

We describe the details of our anchor-free detection head that works across views and formats.
The key is to abstract away from the specific views and formats, and think about the individual \emph{elements}. 
The elements are individual voxels / pixels / pillars under the voxel / perspective / pillar view. 

\subsection{Training Phase}

The detection head has two sequential jobs: finding the centers, and regressing parameters from them.

\subsubsection{Finding the Centers}

In training the network to find the centers, we construct a ground truth \emph{heatmap}.
For each element $e \in E$ where $E$ is the set of all elements, we use $V(e)$ to represent its Cartesian coordinates, which can be either 2-dimensional (only $x$ and $y$) or 3-dimensional (all of $x, y, z$).
We construct its ground truth heatmap value to be:
\begin{equation*}
    h(e) = \max_{c \in C(e)} \exp(- \frac{|| V(e) - c || - \min_{f \in E} || V(f) - c ||}{\sigma^2})
\end{equation*}
where $C(e)$ is the set of centers of the boxes that contain $e$, and $\sigma$ is a hyperparameter. 
$h(e) = 0$ if $|C(e)| = 0$.
Intuitively, the heatmap value is high when the element is close to an object center ($|| V(e) - c||$). 
This distance is modified / compensated by the closest distance among all the elements ($\min_{f \in E} || V(f) - c ||$).

A penalty-reduced focal loss~\cite{lin2017focal,zhou2019objects} is used to train the predicted heatmap:
\begin{align*}
    L_{\text{center}} = &-\frac{1}{|E|} \sum_{e \in E} \{ (1 - \tilde{h}(e))^\alpha \log (\tilde{h}(e)) \mathbb{I}_{h(e) > 1 - \epsilon} + \\
    & (1 - h(e))^\beta \tilde{h}(e)^\alpha \log (1 - \tilde{h}(e)) \mathbb{I}_{h(e) \leq 1 - \epsilon} \}
\end{align*}
where $\tilde{h}(e)$ is the predicted heatmap value for element $e$, $\alpha = 2$, $\beta = 4$, $\epsilon = 0.001$.

\subsubsection{Regressing Box Parameters}

We use smooth L1 loss to regress the 3-dimensional box center offsets, as well as the 3-dimensional box length, width, height.
We use a bin loss~\cite{shi2019pointrcnn} to regress the heading. 
We also add a IoU loss~\cite{zhou2019iou}.
These losses are only active for elements that have ground truth heatmap values greater than a threshold $\delta$.

\subsection{Inference Phase}

After the forward pass produces the predicted heatmap $\tilde{h}$, the predicted object centers are the elements whose predicted heatmap value exceeds a threshold \emph{and} is the local maximum.
The latter is achieved by max pooling (possible on both dense grids and sparse) within a local window ($3 \times 3$ or $3 \times 3 \times 3$). 
The box parameters prediction on these elements complete the inference.

We conclude by reiterating that when the view is voxel and the format is sparse, this detection head exactly follows RSN~\cite{sun2021rsn}.

\section{Randomly Generated Architectures}

We describe our procedure of randomly generating architectures stage by stage.

For the first stage, we add each view with probability $0.5$ independently. 
For views that may have either dense or sparse formats, the format is selected with equal probability.
The pillar / voxel size is $0.32$m to avoid voxelization mismatch complications.
The number of channels is $32$ multiplied by either $0.8$ or $1.0$ or $1.2$.
The layer progression is randomly chosen between five choices.
If the layer type is point, this means the number of dense-normalization-ReLU is between $1$ and $5$. 
For the other layer types, this means the number of downsampling / upsampling scales choose between $(0, 0)$, $(1, 0)$, $(2, 0)$, $(2, 1)$, $(2, 2)$.

For the second stage, we again add each view with probability $0.5$ independently.
For each added view, we iterate through the views selected in the first stage, and add it to the ancestor with probability $0.5$ independently.
The generation process for the other parameters (pillar / voxel size, number of channels, layer progression) is the same as the first stage.

The third stage should only contain one view.
We select the view among voxel, perspective, pillar with equal probability. 
For views that may have either dense or sparse formats, the format is selected with equal probability.
For this selected view, all the views selected in the second stage are its ancestors. 
The generation process for the other parameters is the same as the preceding stages.

The randomly generated architecture may be invalid for several reasons.
Examples include: no branches are added in a particular stage; no ancestors are selected for a second stage view; there may be views in the first stage that are not selected by any view in the second stage.
If any of these situations happen, we reject the sample and sample again until it succeeds.

\end{document}